\documentclass[11pt]{article}

\usepackage[final]{acl}

\usepackage{times}
\usepackage{latexsym}
\usepackage{amsmath}
\usepackage{amssymb}
\usepackage{booktabs}
\usepackage{xcolor}
\usepackage{colortbl} 
\usepackage{tcolorbox}
\usepackage{algorithm}
\usepackage{algpseudocode} 
\usepackage{multirow}
\usepackage[T1]{fontenc}

\usepackage[utf8]{inputenc}

\usepackage{microtype}

\usepackage{inconsolata}

\usepackage{graphicx}

\usepackage{subfigure}
\tcbuselibrary{breakable} 

\title{Placing Puzzle Pieces Where They Matter: A Question Augmentation Framework for Reinforcement Learning}

\author{
    Yangyi Fang\textsuperscript{1}, 
    Jiaye Lin\textsuperscript{1}, 
    Xiaoliang Fu\textsuperscript{2}, 
    Cong Qin\textsuperscript{3}, 
    \textbf{Haolin Shi}\textsuperscript{1,$\dagger$}\\
    \textsuperscript{1}Tsinghua University \quad
    \textsuperscript{2}Fudan University \quad
    \textsuperscript{3}Peking University \\
    \texttt{shihaolin0720@gmail.com}
}

\begin{document}

\maketitle

\begingroup
  \renewcommand\thefootnote{}
  \footnotetext{%
    Authors are listed in no particular order. \\[2pt]
    \textsuperscript{$\dagger$}\ Corresponding author.
  }
\endgroup

\begin{abstract}
Reinforcement learning has become a powerful approach for enhancing large language model reasoning, but faces a fundamental dilemma: training on easy problems can cause overfitting and pass@k degradation, while training on hard problems often results in sparse rewards. Recent question augmentation methods address this by prepending partial solutions as hints. However, uniform hint provision may introduce redundant information while missing critical reasoning bottlenecks, and excessive hints can reduce reasoning diversity, causing pass@k degradation. We propose \textbf{PieceHint}, a hint injection framework that strategically identifies and provides critical reasoning steps during training. By scoring the importance of different reasoning steps, selectively allocating hints based on problem difficulty, and progressively withdrawing scaffolding, PieceHint enables models to transition from guided learning to independent reasoning. Experiments on six mathematical reasoning benchmarks show that our 1.5B model achieves comparable average performance to 32B baselines while preserving pass@k diversity across all $k$ values.
\end{abstract}

\section{Introduction}

Reinforcement learning (RL) has emerged as a powerful paradigm for enhancing LLM reasoning capabilities~\cite{guo2025deepseek,moshkov2025aimo}. However, RL-based training faces a fundamental dilemma: training on easy problems causes overfitting and pass@k degradation, while training on hard problems yields sparse reward signals~\cite{yue2025does,shao2024deepseekmath}.

\begin{figure}[t]
\centering
\includegraphics[width=\columnwidth]{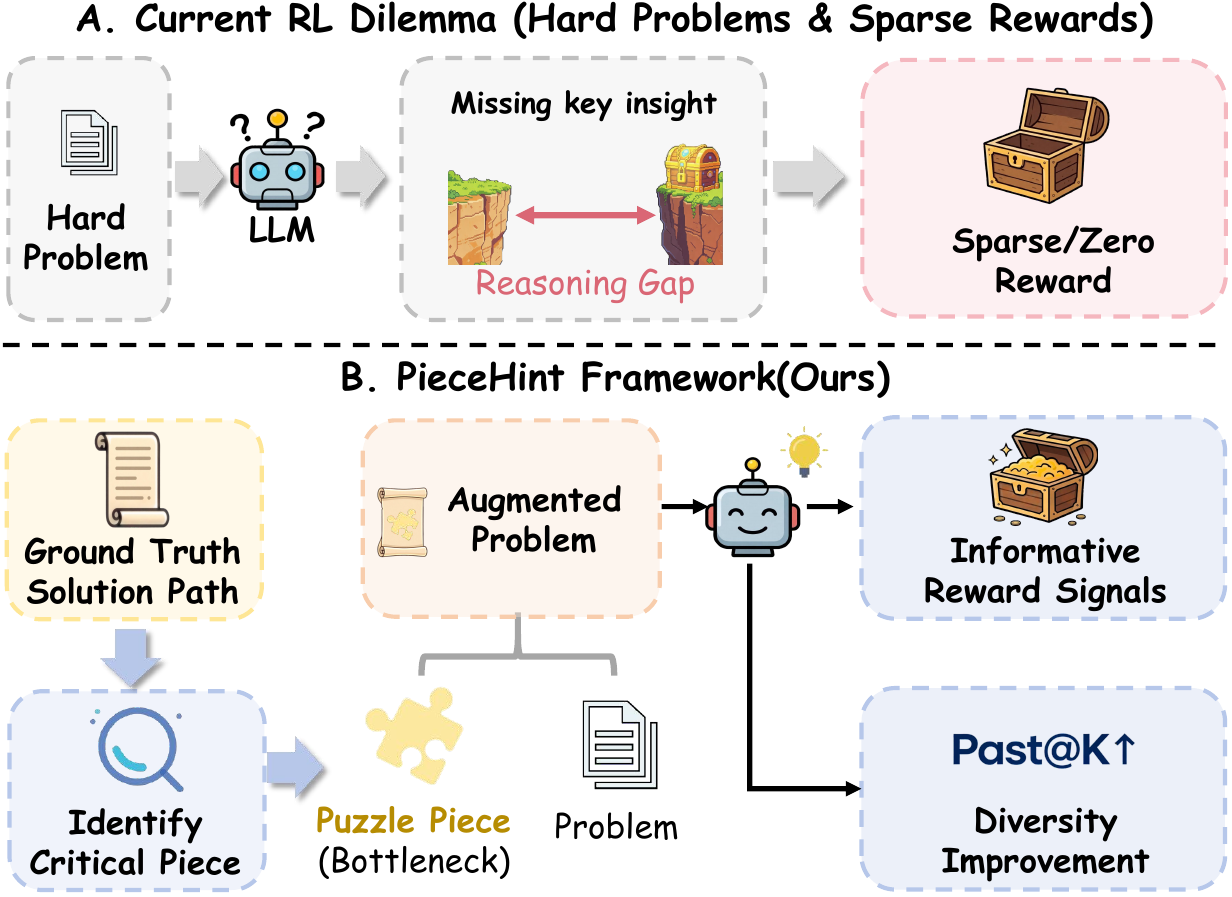}
\caption{RL training dilemma and PieceHint solution. \textbf{Top}: Missing key insights at reasoning bottlenecks lead to sparse/zero rewards. \textbf{Bottom}: PieceHint identifies critical pieces from ground-truth solutions, injects them as hints to bridge gaps, enabling informative rewards and pass@k diversity.}
\label{fig:intro}
\end{figure}

Various approaches address this challenge through reward shaping~\cite{zhu2025surprising,shao2025spurious,wang2026anchoredpolicyoptimizationmitigating, wang2026spposequencelevelppolonghorizon}, process-level supervision~\cite{wang2024mathshepherdverifyreinforcellms,malik2025rewardbench2advancingreward}, and curriculum learning~\cite{deepscaler2025,parashar2025curri,yan2025inftythink,li2025curriculum}. Question augmentation offers a complementary direction: StepHint~\cite{zhang2025stephint} prepends partial ground-truth solutions as hints, enabling models to discover successful trajectories on hard problems. However, uniform hint provision may introduce redundant information while missing critical bottlenecks, and excessive hints can reduce reasoning diversity, causing pass@k degradation.

Figure~\ref{fig:intro} illustrates the core challenge and our solution. In reinforcement learning, models receive rewards only upon generating correct final answers—partial progress yields no credit. When facing hard problems, models often fail at specific reasoning bottlenecks—missing a pivotal lemma, overlooking a non-obvious insight, or lacking a critical technique. These bottlenecks act as missing puzzle pieces: without them, the reasoning chain cannot be completed. If all attempts fail at these bottlenecks, reward variance becomes zero, eliminating gradient signals for learning (Figure~\ref{fig:intro}, top). Uniform hint provision may address routine derivations the model can already handle, while leaving critical reasoning gaps unbridged. This reveals a key insight: reasoning breakthroughs occur at specific bottlenecks rather than uniformly across steps, and identifying and providing these critical puzzle pieces yields maximum learning benefit.

To address this challenge, we propose \textbf{PieceHint}, which treats critical reasoning steps as puzzle pieces that unlock solution pathways (Figure~\ref{fig:intro}, bottom). Our approach strategically identifies high-value pieces through value-driven scoring, allocates them based on problem difficulty and model capability, and progressively withdraws them during training, enabling models to transition from scaffolded learning to independent reasoning.

Our contributions are:
\begin{itemize}
\item We propose PieceHint, a hint injection framework that identifies critical reasoning pieces and strategically provides them during training.

\item We introduce progressive piece withdrawal curriculum, enabling transition from scaffolded to hint-free reasoning.

\item Experiments on six mathematical benchmarks show our 1.5B model achieves comparable average performance to 32B baselines while demonstrating superior parameter efficiency and preserving exploration diversity.
\end{itemize}

\section{Related Works}

\paragraph{Reinforcement Learning for Reasoning.}
Reinforcement learning has emerged as a powerful paradigm for enhancing language model reasoning capabilities, particularly in mathematical problem-solving~\cite{guo2025deepseek,moshkov2025aimo}. Foundational algorithms such as PPO~\cite{schulman2017proximalpolicyoptimizationalgorithms} and GRPO~\cite{guo2025deepseek} establish the algorithmic basis for policy optimization with outcome supervision. Subsequent work has explored orthogonal improvements across multiple dimensions: reward shaping to provide intermediate signals~\cite{zhu2025surprising,shao2025spurious}, process-level supervision to credit individual reasoning steps~\cite{wang2024mathshepherdverifyreinforcellms,malik2025rewardbench2advancingreward}, and chain-of-thought verification to validate reasoning trajectories~\cite{wen2025reinforcement}.

\paragraph{Sample Efficiency and Exploration.}
Parallel efforts address sample efficiency and exploration through uncertainty-aware training strategies. TreeRL~\cite{hou2025treerlllmreinforcementlearning} and VinePPO~\cite{kazemnejad2025vinepporefiningcreditassignment} leverage entropy or confidence metrics to prioritize high-quality rollouts and improve credit assignment. MRT~\cite{qu2025optimizingtesttimecomputemeta} reuses partial trajectories at inference time to amortize computational cost across multiple queries~\cite{zhou2026one}. Curriculum learning strategies have also been proposed to organize training progression, either by solution length~\cite{deepscaler2025}, problem difficulty~\cite{parashar2025curri}, or reasoning complexity~\cite{yan2025inftythink}.

\paragraph{The Dilemma of Problem Difficulty in RL Training.}
Recent work~\cite{shao2024deepseekmath,yue2025does,zhao2025echochamberrlposttraining} reveals a fundamental tension in RL-based reasoning: training on easy problems can lead to pass@k degradation as models may overfit to familiar solution patterns, while training on hard problems often results in sparse reward signals. To mitigate pass@k degradation, several methods~\cite{yu2025dapo,liu2025prorl} employ entropy regularization through adaptive clipping, temperature tuning, or periodic reference model updates. These algorithmic modifications operate at the optimization level, adjusting how models learn from existing rollouts.

\paragraph{Question Augmentation for RL.}
Question augmentation operates at the input level to reshape problem difficulty. StepHint~\cite{zhang2025stephint} prepends sequential partial solutions (e.g., the first 50\% of reasoning steps), bridging the gap between easy and hard problems while improving convergence speed and pass@k preservation. An alternative approach~\cite{yan2025learning} provides entire reasoning chains from stronger models as reference trajectories, though the target model still explores independently most of the time. While these methods demonstrate effectiveness, position-based selection may include routine steps while potentially missing non-obvious insights that appear later in solutions, as validated by our ablation studies (Section~\ref{subsec:ablation}). In contrast to position-based approaches, our method identifies critical reasoning bottlenecks through value-driven piece identification and scoring, strategically providing hints based on their impact on learning rather than sequential position. Furthermore, through progressive piece withdrawal, we enable models to transition from scaffolded learning to hint-free reasoning, addressing potential generalization limitations of fixed augmentation methods.

\begin{figure*}[t]
\centerline{\includegraphics[width=1\textwidth]{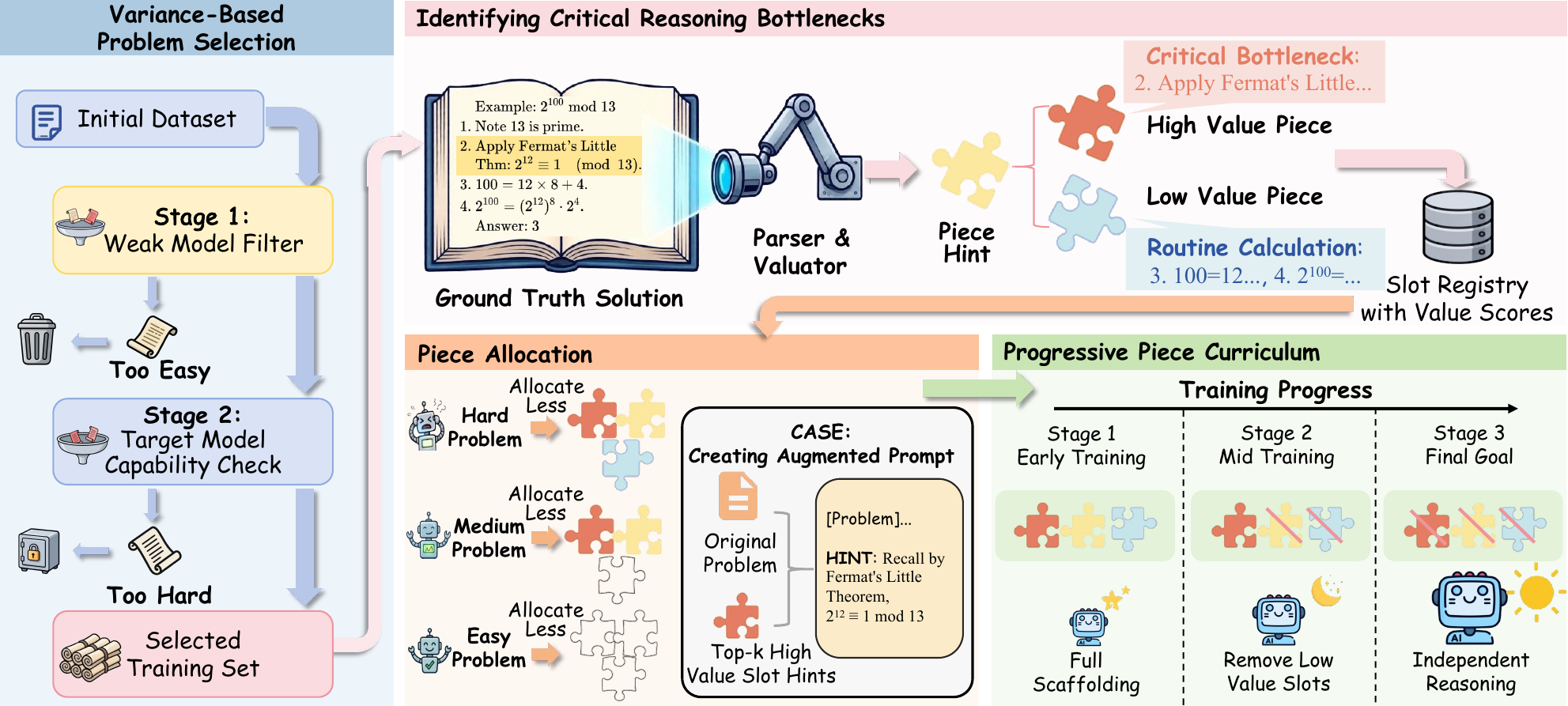}}
\caption{Overview of PieceHint. We identify critical reasoning pieces via value scores, allocate them by model capability, and progressively withdraw hints during training for hint-free reasoning.}
\label{fig:overview}
\end{figure*}

\section{Preliminaries}
\label{sec:preliminaries}

\paragraph{Problem Formulation.}
We consider mathematical reasoning as a sequence generation task. Given a problem $q$ from problem space $\mathcal{Q}$, the goal is to generate a reasoning trajectory $o$ that leads to the correct answer. We denote the ground-truth solution as $\tau^* = [\tau^*_1, \tau^*_2, \ldots, \tau^*_n]$, which is a sequence of reasoning steps, and the final answer as $a^*$. A generated response $o$ is considered correct if its extracted answer matches $a^*$.

In question augmentation, we construct augmented prompts by prepending auxiliary information as hints to the original problem. Formally, an augmented prompt is denoted as $\tilde{q} = [h; q]$, where $h$ represents the hint and $[;]$ denotes concatenation. The hint $h$ can take various forms depending on the augmentation strategy—for instance, partial solution steps, key intermediate results, or critical reasoning components. During RL training, the model learns from augmented prompts $\tilde{q}$ while being evaluated on original prompts $q$ without any hints, ensuring the learned policy generalizes to hint-free scenarios.

\paragraph{Group Relative Policy Optimization.}
We adopt Group Relative Policy Optimization (GRPO)~\citep{shao2024deepseekmath} as our reinforcement learning algorithm. GRPO addresses the high variance problem in policy gradient methods by sampling multiple rollouts per prompt and normalizing advantages within each group.

Formally, for each prompt $q$, GRPO samples a group of $G$ rollouts $\{o_i\}_{i=1}^{G}$ from the current policy $\pi_{\text{old}}$. The advantage function for the $i$-th rollout is computed via group-wise normalization:
\begin{equation}
\small
A_{i,t} = \frac{R(o_i) - \bar{R}}{\sigma_R},
\end{equation}
where $R(o_i)$ denotes the reward for response $o_i$, and $\bar{R}$ and $\sigma_R$ represent the mean and standard deviation of rewards within the group $\{R(o_j)\}_{j=1}^G$, respectively. This advantage $A_{i,t}$ is shared across all tokens in the sequence.

The optimization objective maximizes a clipped surrogate loss:
\begin{equation}
\small
\begin{aligned}
\mathcal{L}(\theta) &= \mathbb{E}_{q \sim \mathcal{D}, \{o_i\} \sim \pi_{\text{old}}} \Bigg[
\frac{1}{\sum_{i} |o_i|} \sum_{i=1}^{G} \sum_{t=1}^{|o_i|} \\
&\quad \min \Big( r_{i,t} A_{i,t}, \text{clip}(r_{i,t}, 1-\epsilon, 1+\epsilon) A_{i,t} \Big)
\Bigg],
\end{aligned}
\end{equation}
where $r_{i,t} = \pi_\theta(o_{i,t} \mid q,o_{i,<t}) / \pi_{\text{old}}(o_{i,t} \mid q,o_{i,<t})$ is the importance sampling ratio, $\epsilon$ is the clipping parameter, and $|o_i|$ denotes the length of the $i$-th response. Following recent work~\citep{yu2025dapo, liu2025understanding}, we omit the KL divergence penalty term in our implementation.

We employ a binary reward function $R: \mathcal{Q} \times \mathcal{V}^* \to \{0, 1\}$, where $R(q, o_i) = 1$ if the final answer is correct and properly formatted, and $R(q, o_i) = 0$ otherwise. This sparse reward design provides clear credit assignment but can lead to insufficient learning signals on extremely difficult problems where success rates approach zero.

\section{Methodology}

The key insight driving our approach is that reasoning steps contribute unequally to problem-solving success: some represent critical bottlenecks—pivotal lemmas, non-obvious insights, or key techniques—where models frequently fail, while others are routine derivations models can handle independently. This motivates PieceHint, a value-driven framework that strategically provides hints at high-impact bottlenecks rather than uniformly covering all steps. 

Our method operates through a pipeline combining offline preprocessing and online training (Figure~\ref{fig:overview}). In the offline phase, we apply variance-based problem selection (Section~\ref{subsec:variance_selection}) to identify problems in the optimal difficulty range from initial dataset $\mathcal{Q}_{\text{init}}$, yielding training set $\mathcal{Q}_{\text{train}}$. For each problem $q \in \mathcal{Q}_{\text{train}}$, value-driven piece identification (Section~\ref{subsec:bottleneck_id}) extracts critical reasoning pieces from ground-truth solutions and assigns importance scores, followed by piece allocation (Section~\ref{subsec:adaptive_allocation}) that determines initial hint budgets based on model capability. During online training, progressive piece curriculum (Section~\ref{subsec:progressive_curriculum}) gradually withdraws hints, enabling transition from scaffolded learning to hint-free reasoning.

\subsection{Variance-Based Problem Selection}
\label{subsec:variance_selection}

Not all problems are equally suitable for reinforcement learning. Trivially easy problems cause overfitting and degrade exploration, while prohibitively difficult problems provide insufficient reward signals~\cite{shao2024deepseekmath,yue2025does}. We propose a two-stage selection strategy identifying problems in the optimal difficulty range—challenging yet learnable.

\noindent\textbf{Identifying hard problems.}
Given initial dataset $\mathcal{Q}_{\text{init}}$, we employ a weak selection model $\mu_{\text{weak}}$ (comparable or slightly weaker than our training model) to filter trivially easy problems. For each problem $q \in \mathcal{Q}_{\text{init}}$, we sample $m$ reasoning attempts from $\mu_{\text{weak}}$, counting correct solutions as $c_{\text{weak}}(q)$, and construct the hard problem subset:
\begin{equation}
\small
\mathcal{Q}_{\text{hard}} = \{q \in \mathcal{Q}_{\text{init}} \mid c_{\text{weak}}(q) \leq \alpha_1 \cdot m\},
\label{eq:hard_filter}
\end{equation}
where $\alpha_1$ is a threshold parameter, retaining problems where $\mu_{\text{weak}}$ succeeds at most $\alpha_1 \cdot m$ times out of $m$ attempts.

\noindent\textbf{Capability-aligned selection.}
Among $\mathcal{Q}_{\text{hard}}$, we select problems aligning with our training model $\mu_0$'s current capability. We evaluate $\mu_0$ on each problem without augmentation, sampling $m$ attempts and computing pass@$m(q, \mu_0) = c_{\text{train}}(q)$ (the number of correct solutions out of $m$ attempts), yielding:
\begin{equation}
\small
\mathcal{Q}_{\text{train}} = \{q \in \mathcal{Q}_{\text{hard}} \mid \alpha_2 \cdot m \leq c_{\text{train}}(q) \leq \alpha_3 \cdot m\},
\label{eq:capability_filter}
\end{equation}
where $\alpha_2$ and $\alpha_3$ bound the acceptable success rate range.

This two-stage filtering removes problems that are too easy ($c_{\text{weak}}(q) > \alpha_1 \cdot m$), intractable ($c_{\text{train}}(q) = 0$), or already accessible ($c_{\text{train}}(q) > \alpha_3 \cdot m$), retaining those with moderate success rates—the ideal regime for effective RL. We preserve the success counts $\{c_{\text{train}}(q)\}_{q \in \mathcal{Q}_{\text{train}}}$ for subsequent piece allocation in Section~\ref{subsec:adaptive_allocation}.

\subsection{Identifying Critical Reasoning Bottlenecks}
\label{subsec:bottleneck_id}

Having identified learnable problems, we next determine which solution parts are most critical. Traditional methods mechanically truncate solution prefixes; we propose a value-driven approach identifying critical bottlenecks where targeted hints provide maximum benefit.

For each problem $q \in \mathcal{Q}_{\text{train}}$ with ground-truth solution $\tau^* = [\tau^*_1, \ldots, \tau^*_n]$ (defined in Section~\ref{sec:preliminaries}), we employ a stronger LLM to parse $\tau^*$ into a set of reasoning pieces $\mathcal{P}(q) = \{p_1, \ldots, p_n\}$. Each piece $p_i$ encapsulates a logically complete reasoning component that contributes to the solution—for instance, recognizing a key mathematical structure, establishing an intermediate lemma, applying a pivotal calculation technique, or making a critical variable substitution. The granularity of parsing is designed to capture semantic reasoning units rather than arbitrary sentence boundaries, ensuring each piece represents a meaningful cognitive step in the problem-solving process.

\noindent\textbf{Value scoring for bottleneck identification.}
We employ a stronger LLM to assign each piece a discrete value score $V(p_i)$ based on three criteria: (1) \textit{novelty}—whether the insight is non-trivial, (2) \textit{difficulty}—whether deriving this step demands mathematical maturity, and (3) \textit{impact}—whether subsequent reasoning critically depends on this result. High-value pieces correspond to critical bottlenecks where hints substantially reduce problem difficulty (scoring prompt details in Appendix~\ref{app:scoring}).

To enable cross-problem comparison, we normalize scores within each problem via min-max normalization:
\begin{equation}
\small
V_{\min} = \min_{p_j \in \mathcal{P}(q)} V(p_j), \quad V_{\max} = \max_{p_j \in \mathcal{P}(q)} V(p_j),
\label{eq:value_minmax}
\end{equation}
\begin{equation}
\small
D(p_i) = \frac{V(p_i) - V_{\min}}{V_{\max} - V_{\min}},
\label{eq:normalize_value}
\end{equation}
yielding normalized values $D(p_i) \in [0, 1]$ with higher values indicating more critical bottlenecks. In practice, we observe that value distributions within problems exhibit clear stratification, with a small subset of high-value pieces (typically 1-3 per problem) standing out as pivotal reasoning steps. We maintain a global registry $\mathcal{R} = \{(q, \mathcal{P}(q), V, D)\}_{q \in \mathcal{Q}_{\text{train}}}$ storing piece structures and values for all training problems.

\subsection{Piece Allocation}
\label{subsec:adaptive_allocation}

Having identified and scored pieces, we determine how many hints each problem receives. Problems vary in difficulty—some require substantial scaffolding, others benefit from minimal hints preserving exploration. We allocate initial piece budgets based on model capability.

Leveraging the success counts $\{c_{\text{train}}(q)\}_{q \in \mathcal{Q}_{\text{train}}}$ from Section~\ref{subsec:variance_selection}, we assign initial piece budgets via a three-tier allocation function. For conciseness, we denote $c(q) = c_{\text{train}}(q)$ as the number of correct solutions out of $m$ attempts. The allocation function is defined as:
\begin{equation}
\small
k^{(0)}(q) = 
\begin{cases}
k_{\max} & \text{if } c(q) \in [0, \beta_1 \cdot m], \\
\lfloor k_{\max}/2 \rfloor & \text{if } c(q) \in (\beta_1 \cdot m, \beta_2 \cdot m], \\
0 & \text{if } c(q) \in (\beta_2 \cdot m, m],
\end{cases}
\label{eq:tier_allocation}
\end{equation}
where $k^{(0)}(q)$ denotes the initial number of pieces allocated to problem $q$, and $k_{\max}, \beta_1, \beta_2$ are hyperparameters controlling the allocation. Specifically, hard problems ($c(q) \in [0, \beta_1 \cdot m]$) receive maximum support ($k_{\max}$ pieces) enabling successful trajectories; medium problems ($c(q) \in (\beta_1 \cdot m, \beta_2 \cdot m]$) receive moderate support ($\lfloor k_{\max}/2 \rfloor$ pieces) improving consistency; easy problems ($c(q) \in (\beta_2 \cdot m, m]$) receive no hints preserving exploration.

\noindent\textbf{Value-driven piece selection.}
For problems with $k^{(0)}(q) > 0$, we select the top-$k^{(0)}(q)$ pieces maximizing the sum of normalized values:
\begin{equation}
\small
H_{k^{(0)}}(q) = \arg\max_{\substack{S \subseteq \mathcal{P}(q) \\ |S| = k^{(0)}(q)}} \sum_{p_i \in S} D(p_i).
\label{eq:topk_selection}
\end{equation}
This yields an augmented training set $\{(q, H_{k^{(0)}}(q))\}_{q \in \mathcal{Q}_{\text{train}}}$, where $H_{k^{(0)}}(q)$ denotes the initial hint set for problem $q$. During training, pieces in $H_{k^{(0)}}(q)$ are prepended to the original problem to construct augmented prompts $\tilde{q} = [H_{k^{(0)}}(q); q]$.

\subsection{Progressive Piece Curriculum}
\label{subsec:progressive_curriculum}

Having constructed augmented prompts with hints, we ensure the model eventually learns to solve problems independently. Fixed augmentation perpetually providing hints may hinder generalization; we implement a progressive curriculum gradually withdrawing hints during training, enabling transition from scaffolded learning to independent reasoning.

Using the augmented training set $\{(q, H_{k^{(0)}}(q))\}_{q \in \mathcal{Q}_{\text{train}}}$ from Section~\ref{subsec:adaptive_allocation}, we train via GRPO while implementing this curriculum. At each training iteration, we sample problems from $\mathcal{Q}_{\text{train}}$ and construct augmented prompts by prepending the current hint set $H_k(q)$ (initially set to $H_{k^{(0)}}(q)$). The model generates responses and receives binary rewards per Section~\ref{sec:preliminaries}, with policy updates following the GRPO objective. As training progresses, different problems advance through the curriculum at different rates depending on their sampling frequency. Each problem $q$ maintains a sampling counter $s(q)$ (initialized to zero) tracking its training exposure.

\noindent\textbf{Frequency-based withdrawal mechanism.}
When $s(q) \bmod N_{\text{check}} = 0$ (i.e., every $N_{\text{check}}$ samples of problem $q$) and the current hint set is non-empty ($|H_k(q)| > 0$), we remove the least valuable piece to preserve critical bottlenecks longer:
\begin{equation}
\small
\begin{aligned}
p_{\text{remove}} &= \arg\min_{p_i \in H_k(q)} D(p_i), \\
H_k(q) &\leftarrow H_k(q) \setminus \{p_{\text{remove}}\},
\end{aligned}
\label{eq:removal_strategy}
\end{equation}
where $D(p_i)$ is the normalized value from Eq.~\ref{eq:normalize_value}. This implements three design principles: (1) per-problem sampling frequency ensures proportional progression regardless of global training dynamics, (2) removing one piece per $N_{\text{check}}$ samples enables smooth transitions—a problem with $k^{(0)}(q)$ pieces receives $k^{(0)}(q) \times N_{\text{check}}$ training samples with gradually decreasing support, and (3) removing low-value pieces first preserves support at critical bottlenecks while building independence on simpler steps.

For a problem initially assigned $k^{(0)}(q)$ pieces, all hints are withdrawn after exactly $k^{(0)}(q) \times N_{\text{check}}$ samples of that specific problem. Once $H_k(q) = \emptyset$, the problem continues training without any hints, forcing the model to solve it independently. This ensures the model develops robust hint-free reasoning capabilities that generalize to evaluation scenarios where no ground-truth information is available.


\begin{table*}[t]
\small
\centering
{
\setlength{\tabcolsep}{3.5pt}
\renewcommand{\arraystretch}{1.2}
\caption{Main results on mathematical reasoning benchmarks. We report pass@1 accuracy (\%). The best results are \textbf{bolded}, and the second-best results are \underline{underlined}.}
\label{tab:main_results}
\begin{tabular}{lccccccc}
\toprule
\textbf{Method} & \textbf{AIME24} & \textbf{AIME25} & \textbf{AMC23} & \textbf{MATH500} & \textbf{Minerva} & \textbf{Olympiad} & \textbf{Avg.} \\
\midrule
\multicolumn{8}{>{\columncolor{blue!12}}c}{\textit{\textbf{Baseline Models}}} \\
\midrule
\multicolumn{8}{>{\columncolor{blue!8}}l}{\textit{1.5B Models}} \\
DeepSeek-R1-Distill-1.5B~\cite{guo2025deepseek} & 24.4 & 21.8 & 63.6 & 72.1 & 26.2 & 43.4 & 41.9 \\
Qwen3-1.7B~\cite{yang2025qwen3} & 30.6 & 47.5 & 69.0 & 84.1 & 35.4 & 53.7 & 53.4 \\
Nemotron-1.5B~\cite{moshkov2025aimo} & 44.0 & 34.2 & 82.7 & 88.5 & 22.3 & 61.2 & 55.5 \\
\midrule
\multicolumn{8}{>{\columncolor{orange!8}}l}{\textit{4-8B Models}} \\
DeepSeek-R1-Distill-7B~\cite{guo2025deepseek} & 39.6 & 29.9 & 78.8 & 80.3 & 40.0 & 55.1 & 54.0 \\
Qwen3-4B~\cite{yang2025qwen3} & 47.5 & 37.3 & 80.0 & 85.1 & \underline{40.4} & 58.5 & 58.1 \\
\midrule
\multicolumn{8}{>{\columncolor{purple!8}}l}{\textit{32B Models}} \\
DeepSeek-R1-Distill-32B~\cite{guo2025deepseek} & \underline{53.1} & \underline{39.5} & \underline{87.3} & \underline{86.0} & \textbf{45.1} & \underline{61.2} & \underline{62.0} \\
\midrule
\multicolumn{8}{>{\columncolor{teal!15}}c}{\textit{\textbf{PieceHint (Ours)}}} \\
\midrule
\rowcolor{teal!8} PieceHint-DeepSeek-1.5B & 35.8 & 25.5 & 71.2 & 87.7 & 36.7 & 50.1 & 51.2 \\
\rowcolor{teal!8} PieceHint-Qwen3-1.7B & 42.2 & 38.8 & 80.4 & 89.8 & 41.6 & 62.4 & 59.2 \\
\rowcolor{teal!8} \textbf{PieceHint-Nemotron-1.5B} & \textcolor{teal}{\textbf{54.5}} & \textcolor{teal}{\textbf{43.7}} & \textcolor{teal}{\textbf{89.1}} & \textcolor{teal}{\textbf{91.3}} & 28.5 & \textcolor{teal}{\textbf{68.3}} & \textcolor{teal}{\textbf{62.6}} \\
\bottomrule
\end{tabular}
}
\end{table*}

\section{Experiments}

\subsection{Experimental Setup}

\paragraph{Datasets.}
We construct training datasets through a systematic filtering pipeline applied to the OpenR1-Math-220K dataset. Following the variance-based problem selection strategy described in Section~\ref{subsec:variance_selection}, we employ DeepSeek-R1-Distill-1.5B as the weak selection model $\mu_{\text{weak}}$ to filter trivially easy problems. Each base model then uses its own initial checkpoint as the capability reference model $\mu_0$ to perform capability-aligned selection, resulting in model-specific training sets $\mathcal{Q}_{\text{train}}$ containing problems within the target difficulty range for reinforcement learning.

\paragraph{Implementation Details.}
Our implementation is based on the VeRL training framework~\cite{sheng2025hybridflow}. We configure the training pipeline with a rollout batch size of 512 and an update batch size of 32, performing 16 rollout samples per prompt. During rollout generation, we employ nucleus sampling with Top-p $p=1.0$ and temperature $\tau=1.0$. We train for 2,400 gradient update steps, corresponding to approximately 1 million rollout samples. All experiments are conducted on 8$\times$NVIDIA A100 80GB GPUs. Full hyperparameters are provided in Appendix~\ref{app:training_config}.

\paragraph{Evaluation.}
We evaluate our method on six mathematical reasoning benchmarks: AIME24~\cite{AIME}, AIME25~\cite{AIME}, AMC23~\cite{AMC}, MATH500~\cite{hendrycks2021measuring}, Minerva~\cite{lewkowycz2022solving}, and Olympiad~\cite{he2024olympiadbench}. During evaluation, we use Top-p sampling with $p=1.0$ and temperature $\tau=1.0$. We report pass@$k$ metrics following~\cite{RLLimit} and~\cite{unbiasedpassk}, with answer verification performed using the Math-Verify tool~\cite{mathverify}.

\subsection{Main Results}

Table~\ref{tab:main_results} compares PieceHint against baseline models across six mathematical reasoning benchmarks. PieceHint-Nemotron-1.5B achieves competitive performance with the 32B DeepSeek baseline, reaching comparable average accuracy while using 20× fewer parameters (1.5B vs. 32B), with particularly strong performance on competition-level problems (AIME24, AIME25, AMC23) and MATH500. PieceHint-Qwen3-1.7B and PieceHint-DeepSeek-1.5B also demonstrate substantial improvements over their respective base models, outperforming significantly larger 4B baselines. By strategically identifying critical reasoning bottlenecks through value-driven piece selection, allocating hints based on problem difficulty, and progressively withdrawing scaffolding during training, small models can develop robust reasoning capabilities that rival those achieved through conventional parameter scaling.


\begin{table*}[t]
\small
\centering
{
\setlength{\tabcolsep}{2.8pt}
\renewcommand{\arraystretch}{1.15}
\caption{Comprehensive ablation studies on PieceHint components. All experiments use Nemotron-1.5B as the base model. The best result in each group is \textbf{bolded} with distinct colors.}
\label{tab:ablation_unified}
\begin{tabular}{llcccccccc}
\toprule
\textbf{Component} & \textbf{Method} & \textbf{AIME24} & \textbf{AIME25} & \textbf{AMC23} & \textbf{MATH500} & \textbf{Minerva} & \textbf{Olympiad} & \textbf{Avg.} \\
\midrule
\multicolumn{9}{>{\cellcolor[HTML]{FFDADE}}c}{\textit{\textbf{Problem Selection Strategy}}} \\
\multirow{3}{*}{\parbox{2cm}{\centering Problem\\Selection}} 
& Hard Problems Only & 49.8 & 39.2 & 84.6 & 88.1 & 26.3 & 63.5 & 58.6 \\
& Random Selection & 52.9 & 42.1 & 87.8 & 90.2 & 27.8 & 66.8 & 61.3 \\
& \textbf{Variance-Based} & \textcolor{red!50}{\textbf{54.5}} & \textcolor{red!50}{\textbf{43.7}} & \textcolor{red!50}{\textbf{89.1}} & \textcolor{red!50}{\textbf{91.3}} & \textcolor{red!50}{\textbf{28.5}} & \textcolor{red!50}{\textbf{68.3}} & \textcolor{red!50}{\textbf{62.6}} \\
\midrule
\multicolumn{9}{>{\cellcolor[HTML]{C7F0F8}}c}{\textit{\textbf{Piece Selection Strategy}}} \\
\multirow{4}{*}{\parbox{2cm}{\centering Piece\\Selection}} 
& 50\% Prefix Truncation & 46.8 & 36.2 & 82.5 & 86.3 & 25.8 & 61.4 & 56.5 \\
& Random Piece Selection & 49.2 & 38.9 & 85.1 & 88.2 & 26.7 & 63.8 & 58.7 \\
& 25\% Prefix Truncation & 51.7 & 41.3 & 86.8 & 89.5 & 27.3 & 65.9 & 60.4 \\
& \textbf{Value-Driven} & \textcolor{cyan!50}{\textbf{54.5}} & \textcolor{cyan!50}{\textbf{43.7}} & \textcolor{cyan!50}{\textbf{89.1}} & \textcolor{cyan!50}{\textbf{91.3}} & \textcolor{cyan!50}{\textbf{28.5}} & \textcolor{cyan!50}{\textbf{68.3}} & \textcolor{cyan!50}{\textbf{62.6}} \\
\midrule
\multicolumn{9}{>{\cellcolor[HTML]{FFEBDE}}c}{\textit{\textbf{Progressive Curriculum}}} \\
\multirow{2}{*}{\parbox{2cm}{\centering Progressive\\Curriculum}} 
& w/o Withdrawal & 52.1 & 41.5 & 87.3 & 89.8 & 27.6 & 66.2 & 60.8 \\
& \textbf{w/ Withdrawal} & \textcolor{orange!50}{\textbf{54.5}} & \textcolor{orange!50}{\textbf{43.7}} & \textcolor{orange!50}{\textbf{89.1}} & \textcolor{orange!50}{\textbf{91.3}} & \textcolor{orange!50}{\textbf{28.5}} & \textcolor{orange!50}{\textbf{68.3}} & \textcolor{orange!50}{\textbf{62.6}} \\
\bottomrule
\end{tabular}
}
\end{table*}

\begin{figure*}[t]
\centering
\subfigure[Problem Selection Strategy]{
\begin{minipage}[t]{0.32\linewidth}
\centerline{\includegraphics[width=1\hsize]{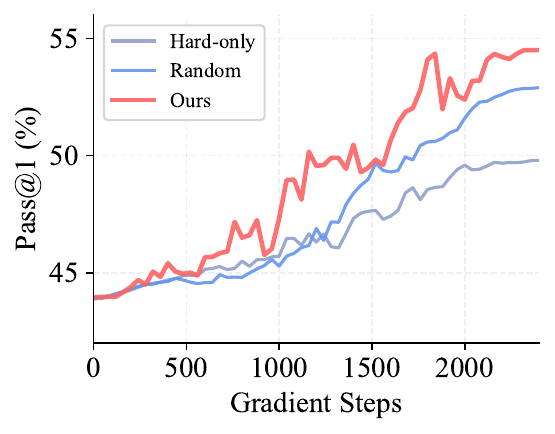}}
\label{fig:dynamics_problem}
\end{minipage}}
\subfigure[Piece Selection Strategy]{
\begin{minipage}[t]{0.32\linewidth}
\centerline{\includegraphics[width=1\hsize]{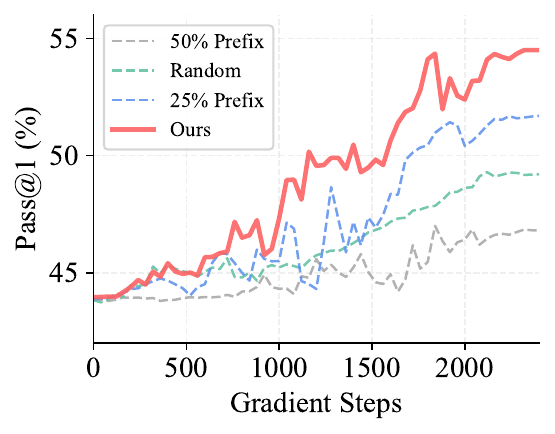}}
\label{fig:dynamics_piece}
\end{minipage}}
\subfigure[Progressive Curriculum]{
\begin{minipage}[t]{0.32\linewidth}
\centerline{\includegraphics[width=1\hsize]{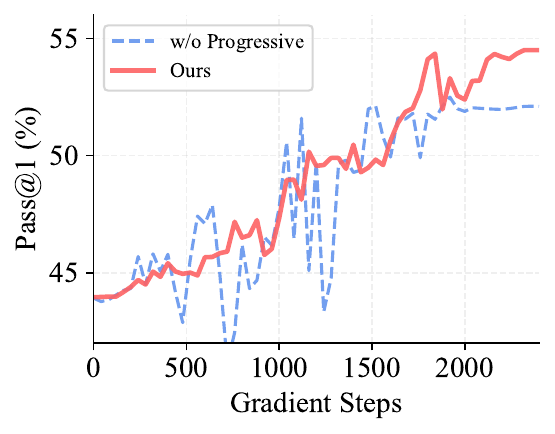}}
\label{fig:dynamics_curriculum}
\end{minipage}}
\caption{Training dynamics on AIME24 across ablation variants. All curves start from similar initial performance and diverge as training progresses, with PieceHint (red solid lines) consistently achieving the highest final performance while maintaining stable convergence.}
\label{fig:training_dynamics}
\end{figure*}

\begin{figure*}[t]
\centering
\subfigure[Nemotron-1.5B]{
\begin{minipage}[t]{0.32\linewidth}
\centerline{\includegraphics[width=1\hsize]{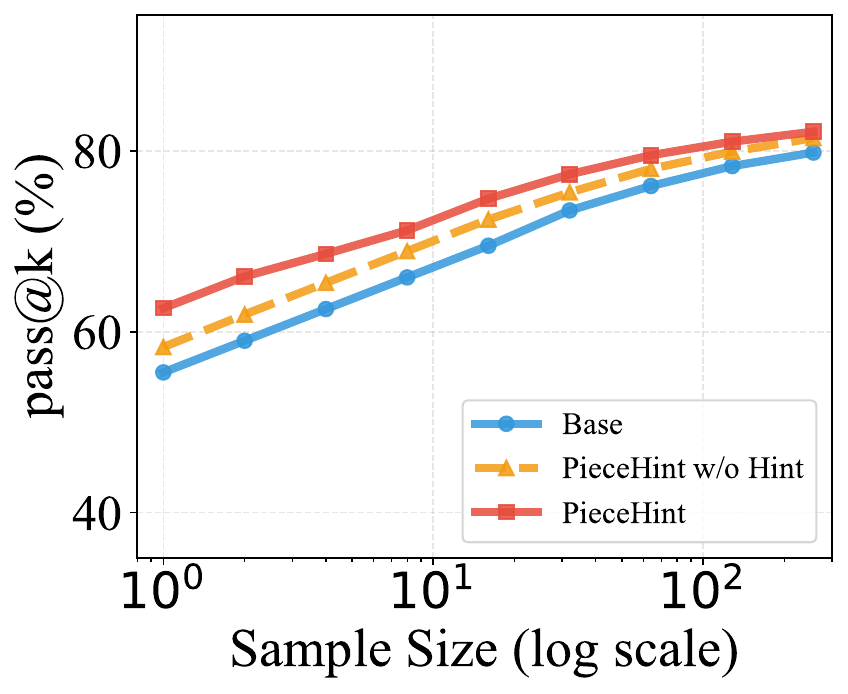}}
\vspace{4pt} 
\label{fig:passk_nemotron}
\end{minipage}}
\subfigure[Qwen3-1.7B]{
\begin{minipage}[t]{0.32\linewidth}
\centerline{\includegraphics[width=1\hsize]{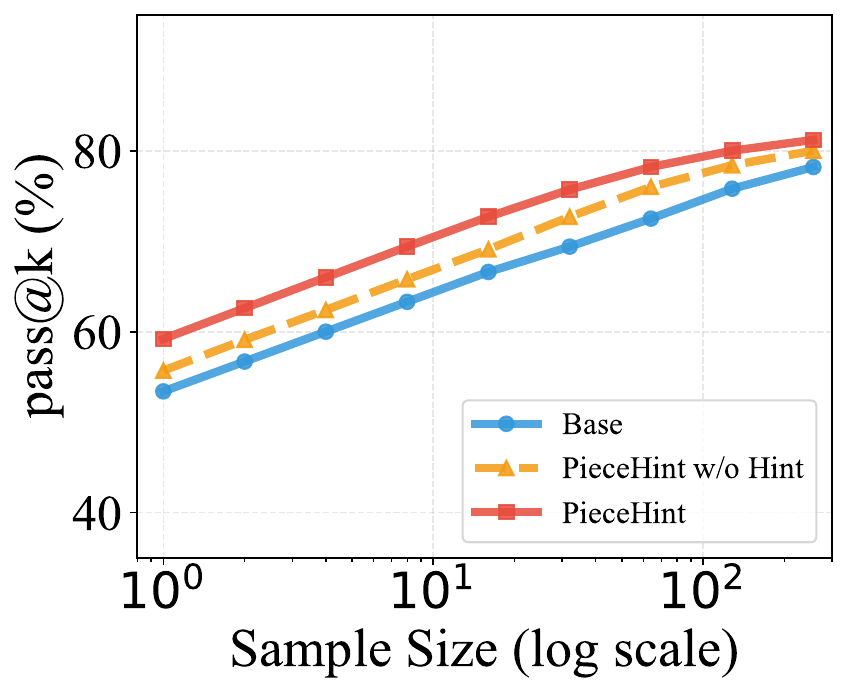}}
\vspace{4pt} 
\label{fig:passk_qwen3}
\end{minipage}}
\subfigure[DeepSeek-R1-Distill-1.5B]{
\begin{minipage}[t]{0.32\linewidth}
\centerline{\includegraphics[width=1\hsize]{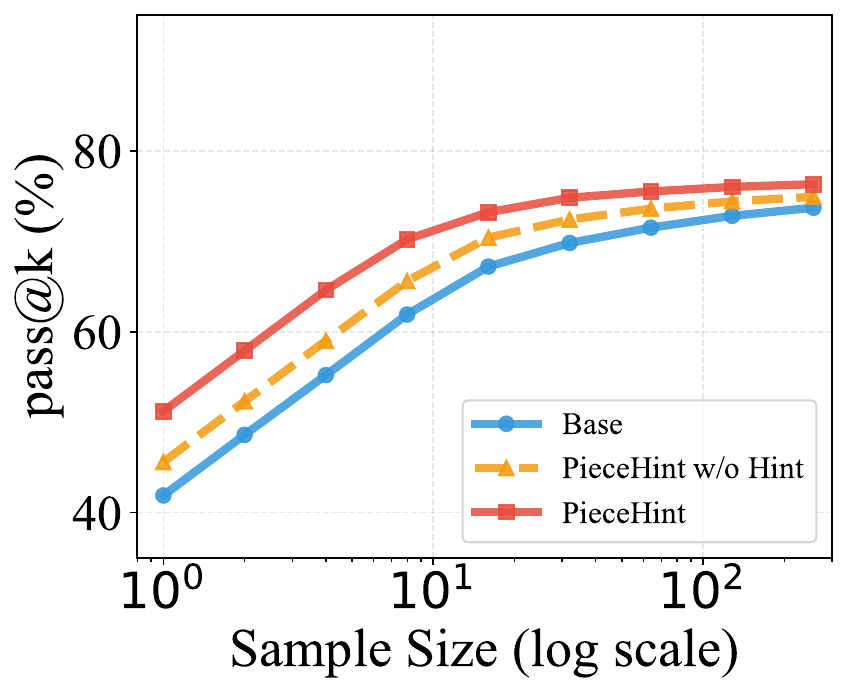}}
\vspace{4pt} 
\label{fig:passk_deepseek}
\end{minipage}}
\caption{Pass@k performance across different sampling budgets showing convergent behavior at higher k values.}
\label{fig:passk_analysis}
\end{figure*}
\subsection{Ablation Studies}
~\label{subsec:ablation}

We conduct ablation studies to validate the effectiveness of each component in PieceHint. We systematically evaluate three core design choices: variance-based problem selection, value-driven piece selection, and progressive curriculum. All ablation experiments use Nemotron-1.5B as the base model to ensure controlled comparison. Table~\ref{tab:ablation_unified} presents quantitative results, and Figure~\ref{fig:training_dynamics} illustrates the training dynamics.

\paragraph{Impact of Problem Selection Strategy.}
Table~\ref{tab:ablation_unified} (top) and Figure~\ref{fig:dynamics_problem} compare three strategies. Hard-only includes intractable problems providing insufficient signals; random sampling achieves moderate results but lacks systematic difficulty targeting. Variance-based selection outperforms both by focusing on challenging yet learnable problems within the model's capability range, validating that capability-aligned filtering is essential.

\paragraph{Impact of Value-Driven Piece Selection.}
Table~\ref{tab:ablation_unified} (middle) and Figure~\ref{fig:dynamics_piece} compare value-driven piece identification against position-based and random selection baselines. Random selection samples pieces without considering their importance, while prefix truncation strategies select pieces based on sequential position. Among position-based methods, 50\% prefix truncation performs worst by including excessive redundant early steps, while the 25\% variant achieves better results but both miss critical bottlenecks appearing later. Random selection achieves moderate performance but lacks systematic targeting. Value-driven selection consistently outperforms all baselines by strategically identifying pivotal lemmas and key insights regardless of position, demonstrating that bottleneck-focused augmentation is substantially more effective than uniform coverage.

\paragraph{Impact of Progressive Curriculum.}
Table~\ref{tab:ablation_unified} (bottom section) and Figure~\ref{fig:dynamics_curriculum} validate the effectiveness of progressive piece withdrawal. The variant without progressive curriculum maintains fixed hint sets throughout training, leading to over-reliance on scaffolding and degraded performance when evaluated without hints. Progressive withdrawal achieves superior performance by gradually reducing hint dependency while preserving strategic guidance, enabling the model to internalize critical reasoning patterns and generalize robustly to hint-free scenarios.

\subsection{Pass@k Analysis}
\label{subsec:passk}
Figure~\ref{fig:passk_analysis} shows pass@k performance across sampling budgets. PieceHint consistently outperforms both base models and standard RL training across all k values, with particularly pronounced gains at lower k values. The consistent improvement across increasing k demonstrates that strategic piece injection enhances solution diversity rather than creating hint dependency—by providing learning signals at critical bottlenecks and then progressively withdrawing guidance, models develop diverse reasoning strategies through independent exploration. As k increases, the performance gap gradually narrows, though PieceHint maintains advantages throughout the sampling range.

\subsection{Case Study}

\begin{figure}[t]
\centering
\includegraphics[width=\columnwidth]{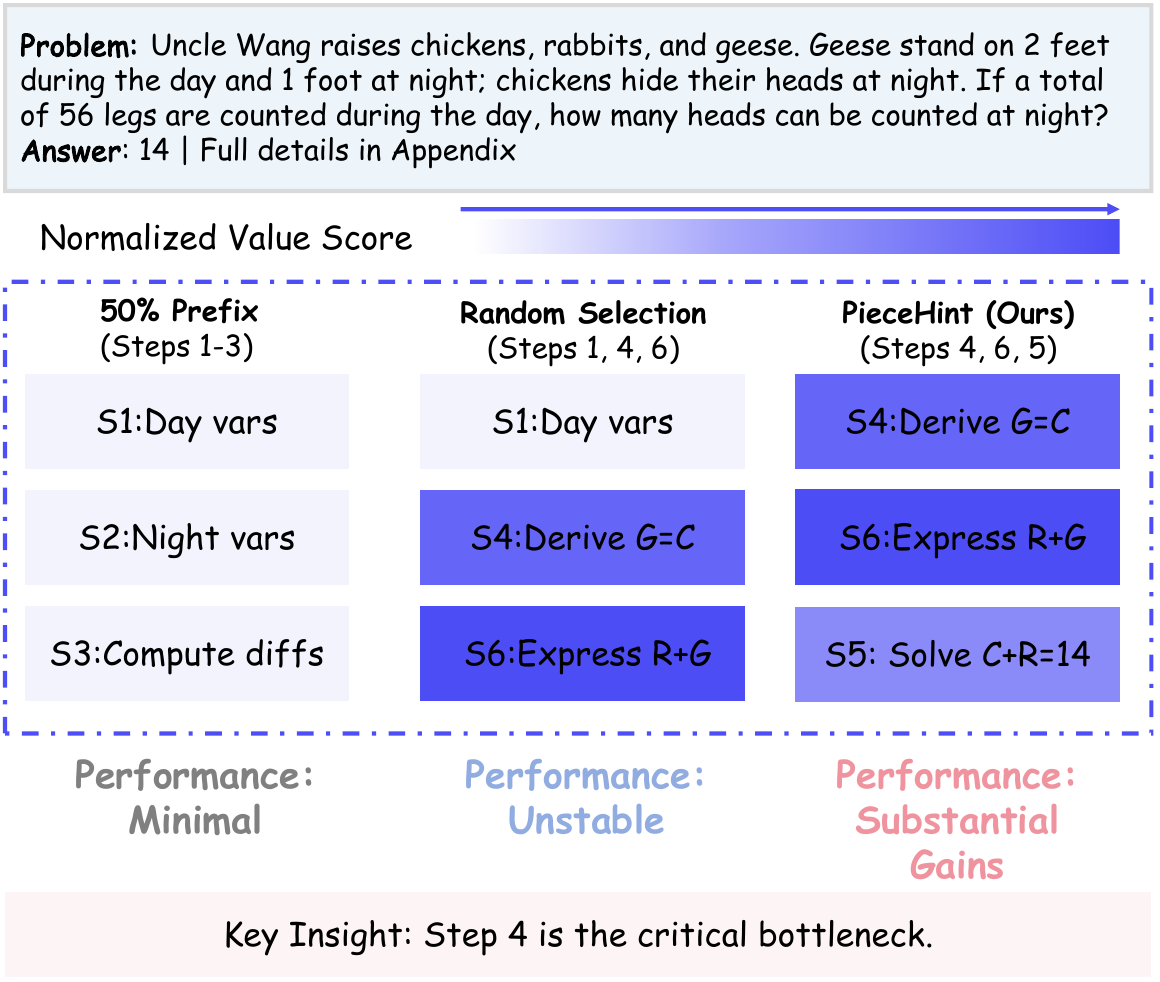}
\caption{Hint strategy comparison (Appendix~\ref{app:case_problem}). PieceHint targets actual learning bottlenecks, unlocking reward signals.}
\label{fig:case_study}
\end{figure}

Figure~\ref{fig:case_study} analyzes hint strategies on a combinatorial problem (Appendix~\ref{app:case_problem}). When models fail at critical reasoning bottlenecks, sparse rewards provide no learning signals. We compare: 50\% prefix wastes hints on routine steps the model can handle; random selection hits bottlenecks inefficiently; PieceHint identifies where models actually struggle and strategically bridges these gaps. This demonstrates that effective augmentation requires understanding which specific reasoning barriers block learning, not just providing arbitrary solution fragments.

\section{Conclusion}

We present PieceHint, a value-driven framework that strategically provides hints at critical reasoning bottlenecks during reinforcement learning. Unlike uniform hint provision, PieceHint identifies where models actually struggle and bridges these gaps through value-driven piece selection, adaptive allocation, and progressive withdrawal, enabling transition from scaffolded to hint-free reasoning. Experiments on six mathematical benchmarks show our 1.5B model achieves comparable performance to 32B baselines while preserving solution diversity, demonstrating remarkable parameter efficiency. We further validate scalability to 7B models (Appendix~\ref{app:scalability}) and generalization to code generation (Appendix~\ref{app:code_generation}), confirming that the core mechanism transfers across model scales and reasoning domains. Future work includes investigating fully adaptive withdrawal strategies that dynamically adjust based on real-time model capability assessments.

\section*{Limitations}
PieceHint demonstrates strong performance across 1.5B and 7B parameter models (Appendix~\ref{app:scalability}), and generalizes to code generation tasks beyond mathematics (Appendix~\ref{app:code_generation}). Our frequency-based withdrawal uses a fixed schedule ($N_{\text{check}}$), trading adaptivity for simplicity. While adaptive strategies that dynamically adjust withdrawal rates and hint allocations based on real-time model capability could potentially improve performance, they would require continuous evaluation of model mastery on each problem to determine optimal training data and hint provision, significantly increasing engineering complexity and computational cost. Our fixed-schedule approach represents a practical trade-off between performance optimization and training efficiency. Additionally, PieceHint's offline preprocessing relies on a stronger LLM as an external oracle; as shown in Appendix~\ref{app:scoring_robustness}, the framework is robust to the choice of scoring model, but this dependency represents a structural boundary condition of the method.

\bibliography{custom}

\appendix

\section{Additional Related Work}
\label{app:related_work}

This section surveys concurrent and complementary work in reinforcement learning for LLM reasoning that further contextualizes PieceHint.

\noindent\textbf{Policy optimization and gradient stability.}
A rich body of recent work improves the optimization dynamics of RL-based LLM training.
Several studies identify structural issues in standard advantage estimation: \citet{yang2026your} analyze biases inherent in group-relative advantage computation, while \citet{yu2026unveiling} reveal implicit symmetry problems in GRPO that hinder exploration and difficulty adaptation—a challenge that PieceHint addresses from the input side by providing targeted scaffolding rather than relying solely on the optimizer.
At the gradient level, \citet{fu2026boldsymbol} address divergence in soft clipping objectives via bilateral decoupled decay of probability gradient weights, and \citet{fu2026maspo} unify gradient utilization, probability mass management, and signal reliability for more robust reasoning.
Complementary to these optimization-level methods, \citet{tan2025bottom} propose bottom-up policy optimization that leverages internal policies latent within pre-trained language models.
Dynamic resource allocation has also been explored: \citet{fang2026allocate} study how to jointly optimize rollout allocation and advantage modulation, which is orthogonal to PieceHint's input-level hint allocation and could be combined to further improve sample efficiency.
These approaches all operate at the optimization level and are compatible with PieceHint as drop-in enhancements to the training pipeline.

\noindent\textbf{Reward modeling and credit assignment.}
Reliable reward signals are critical for RL training.
\citet{cai2025reinforcement} study RL under verifiable yet noisy rewards from imperfect verifiers, and \citet{cai2026vi} propose confidence-guided variance reduction to stabilize verifier-independent RL reasoning—both addressing the signal quality issues that motivate PieceHint's use of binary outcome rewards with strategic hint injection.
\citet{ding2026prpo} propose aligning process-level and outcome-level rewards in policy optimization, offering a supervision signal complementary to PieceHint's piece-level scaffolding.
\citet{liu2026cdrrm} develop contrast-driven rubric generation for more reliable and interpretable reward modeling, and \citet{huang2026does} investigate whether reasoning models can implicitly learn to regulate their own thinking depth—both relevant to understanding the interaction between hint provision and autonomous reasoning development.
Beyond single-turn settings, \citet{fang2026proximity} address credit assignment in multi-turn LLM agent training, sharing the core motivation with PieceHint that sparse terminal rewards are insufficient for effective learning.
On the preference alignment side, \citet{huang2025adaptive} study adaptive sample scheduling for direct preference optimization, and \citet{Huang2026RealTimeAR} explore reward modeling beyond semantic alignment.

\noindent\textbf{Exploration, temperature control, and model efficiency.}
Effective exploration is essential when reward signals are sparse.
\citet{zhou2026look} propose learning temperature policies from LLM internal states via hierarchical RL, dynamically adapting sampling diversity in a manner complementary to PieceHint's progressive hint withdrawal.
\citet{zhang2026heterogeneous} study collaborative RL among heterogeneous agents, extending the multi-agent perspective to settings where diverse reasoning strategies must be coordinated.
On model efficiency, \citet{zhou2025dropping} propose retraining-free pruning for sparse mixture-of-experts models, enabling deployment of capable base models at reduced cost.
Finally, \citet{guo2025tree} apply tree-structured dialogue policy optimization to red-teaming, demonstrating the breadth of domains where RL-based policy learning continues to advance.

\section{Training Configuration Details}
\label{app:training_config}

Tables~\ref{tab:training_config} and~\ref{tab:model_specific} present the complete training configurations. Table~\ref{tab:training_config} shows general hyperparameters shared across all experiments, while Table~\ref{tab:model_specific} presents model-specific problem selection and piece allocation parameters.

\begin{table*}[t]
\centering
\setlength{\tabcolsep}{10pt}
\renewcommand{\arraystretch}{1.15}
\caption{General training hyperparameters and system configuration (shared across all models).}
\label{tab:training_config}
\begin{tabular}{ll|ll}
\toprule
\multicolumn{4}{c}{\textbf{PieceHint-Specific Parameters}} \\
\midrule
Withdrawal Checkpoint ($N_{\text{check}}$) & 2 & Max Initial Pieces ($k_{\max}$) & 3 \\
\midrule
\multicolumn{4}{c}{\textbf{Model Configuration}} \\
\midrule
Gradient Checkpointing & Enabled & Activation Offload & Disabled \\
LoRA Rank & 0 (Disabled) & Fused Kernels & Disabled \\
Remove Padding & Enabled & FSDP Strategy & Fully Sharded \\
Forward Prefetch & Disabled & Reshard After Forward & Enabled \\
Parameter Offload & Disabled & & \\
\midrule
\multicolumn{4}{c}{\textbf{Optimization}} \\
\midrule
Learning Rate & $1 \times 10^{-6}$ & Weight Decay & 0.1 \\
Gradient Clipping & 1.0 & LR Warmup Steps & 10 \\
Warmup Style & Constant & Min LR Ratio & 0.0 \\
\midrule
\multicolumn{4}{c}{\textbf{GRPO Configuration}} \\
\midrule
Mini Batch Size & 32 & Clip Ratio & 0.2 \\
KL Loss Coefficient & 0.0 & Advantage Estimator & GRPO \\
\midrule
\multicolumn{4}{c}{\textbf{Rollout \& Generation}} \\
\midrule
Inference Engine & vLLM & Tensor Parallel Size & 2 \\
Samples per Prompt ($n$) & 16 & Temperature ($\tau$) & 1.0 \\
Top-p ($p$) & 1.0 & Top-k & -1 (Disabled) \\
Max Prompt Length & 2048 & Max Response Length & 8192 \\
Do Sample & True & Ignore EOS & False \\
GPU Memory Utilization & 0.8 & Max Batched Tokens & 10240 \\
Max Num Sequences & 1024 & Chunked Prefill & Enabled \\
Enforce Eager Mode & True & Load Format & dummy\_dtensor \\
\midrule
\multicolumn{4}{c}{\textbf{Data Configuration}} \\
\midrule
Train Batch Size & 512 & Validation Batch Size & 512 \\
Dataloader Workers & 8 & Shuffle Training Data & True \\
Max Prompt Length & 2048 tokens & Max Response Length & 8192 tokens \\
\midrule
\multicolumn{4}{c}{\textbf{Hardware \& Training Schedule}} \\
\midrule
Number of Nodes & 1 & GPUs per Node & 8 \\
GPU Type & A100 80GB & Total GPUs & 8 \\
Total Gradient Steps & 2,400 & Total Rollout Samples & ~1M \\
\bottomrule
\end{tabular}
\end{table*}

\begin{table*}[t]
\small
\centering
\setlength{\tabcolsep}{6pt}
\renewcommand{\arraystretch}{1.2}
\caption{Model-specific problem selection and piece allocation parameters. Each model uses its own checkpoint as the capability reference model $\mu_0$ for variance-based selection.}
\label{tab:model_specific}
\begin{tabular}{lccc}
\toprule
\textbf{Parameter} & \textbf{PieceHint-DeepSeek-1.5B} & \textbf{PieceHint-Qwen3-1.7B} & \textbf{PieceHint-Nemotron-1.5B} \\
\midrule
\multicolumn{4}{c}{\cellcolor{blue!10}\textit{\textbf{Problem Selection Thresholds (Section 4.1)}}} \\
\midrule
$\alpha_1$ (hard filter) & 0.2 & 0.2 & 0.2 \\
$\alpha_2$ (min capability) & 0.1 & 0.1 & 0.1 \\
$\alpha_3$ (max capability) & 0.4 & 0.4 & 0.4 \\
\midrule
\multicolumn{4}{c}{\cellcolor{green!10}\textit{\textbf{Piece Allocation Thresholds (Section 4.3)}}} \\
\midrule
$\beta_1$ (hard tier boundary) & 0.15 & 0.15 & 0.15 \\
$\beta_2$ (medium tier boundary) & 0.35 & 0.35 & 0.35 \\
\bottomrule
\end{tabular}
\end{table*}
\section{Direct Comparison with StepHint}
\label{app:stephint_comparison}

To isolate PieceHint's gains from the choice of hint strategy, we conduct a controlled head-to-head comparison with StepHint~\cite{zhang2025stephint} under identical experimental conditions: the same base model (Nemotron-1.5B), training dataset, hyperparameters, and computational budget. StepHint prepends the first 25\% of ground-truth solution steps as positional prefix hints—the best-performing variant of StepHint as identified in our ablation studies (Table~\ref{tab:ablation_unified}).

\begin{table*}[t]
\small
\centering
\setlength{\tabcolsep}{3.5pt}
\renewcommand{\arraystretch}{1.2}
\caption{Controlled comparison between PieceHint and StepHint under identical settings. Base model: Nemotron-1.5B.}
\label{tab:stephint_comparison}
\begin{tabular}{lccccccc}
\toprule
\textbf{Method} & \textbf{AIME24} & \textbf{AIME25} & \textbf{AMC23} & \textbf{MATH500} & \textbf{Minerva} & \textbf{Olympiad} & \textbf{Avg.} \\
\midrule
No Hints (baseline) & 44.0 & 34.2 & 82.7 & 88.5 & 22.3 & 61.2 & 55.5 \\
StepHint (25\% prefix) & 49.8 & 39.5 & 85.4 & 88.9 & 25.1 & 64.2 & 58.8 \\
\textbf{PieceHint (ours)} & \textbf{54.5} & \textbf{43.7} & \textbf{89.1} & \textbf{91.3} & \textbf{28.5} & \textbf{68.3} & \textbf{62.6} \\
\bottomrule
\end{tabular}
\end{table*}

Table~\ref{tab:stephint_comparison} shows that PieceHint consistently outperforms StepHint across all six benchmarks, with an average improvement of 3.8 points (62.6 vs.\ 58.8). The gap is most pronounced on competition-level benchmarks (AIME24: +4.7, AIME25: +4.2, AMC23: +3.7), where critical bottlenecks tend to appear later in solutions and are systematically missed by positional prefix selection. These results confirm that value-driven identification of reasoning bottlenecks provides a clear advantage over position-based hint allocation: mechanically truncating the first $k$\% of steps may cover routine early derivations while leaving the pivotal insights—which often appear mid-solution—unscaffolded.

\section{Scoring Robustness Analysis}
\label{app:scoring_robustness}

PieceHint relies on an LLM-based scoring step to assign value scores to reasoning pieces. We analyze two potential sources of instability: the choice of scoring model and sensitivity to prompt phrasing.

\subsection*{Cross-Model Scoring Consistency}

We preprocess 1,000 training problems using three different LLM backends and measure both end-to-end performance and the agreement among top-$k$ piece selections.

\begin{table*}[t]
\small
\centering
\setlength{\tabcolsep}{3.5pt}
\renewcommand{\arraystretch}{1.2}
\caption{Performance across different LLM scoring backends (Base model: Nemotron-1.5B).}
\label{tab:scoring_model}
\begin{tabular}{lccccccc}
\toprule
\textbf{Scoring Model} & \textbf{AIME24} & \textbf{AIME25} & \textbf{AMC23} & \textbf{MATH500} & \textbf{Minerva} & \textbf{Olympiad} & \textbf{Avg.} \\
\midrule
GPT-4 (original)         & 54.5 & 43.7 & 89.1 & 91.3 & 28.5 & 68.3 & 62.6 \\
DeepSeek-V3.2            & 54.1 & 43.3 & 88.7 & 91.0 & 28.2 & 67.9 & 62.2 \\
Gemini-3-Flash-Preview   & 53.8 & 43.0 & 88.5 & 90.8 & 28.1 & 67.7 & 62.0 \\
\bottomrule
\end{tabular}
\end{table*}

Performance variance across LLM backends is minimal ($\leq$0.6 average points), confirming that the framework is robust to scoring model choice. To understand why, we further measure cross-scorer agreement on 500 problems:

\begin{table*}[t]
\small
\centering
\setlength{\tabcolsep}{8pt}
\renewcommand{\arraystretch}{1.2}
\caption{Cross-LLM scoring consistency on 500 problems.}
\label{tab:scoring_consistency}
\begin{tabular}{lcc}
\toprule
\textbf{Scorer Pair} & \textbf{Top-3 Piece Overlap} & \textbf{Spearman $\rho$} \\
\midrule
GPT-4 $\leftrightarrow$ DeepSeek-V3.2      & 87\% & 0.86 \\
GPT-4 $\leftrightarrow$ Gemini-3-Flash     & 84\% & 0.83 \\
DeepSeek $\leftrightarrow$ Gemini          & 82\% & 0.80 \\
\bottomrule
\end{tabular}
\end{table*}

High piece-selection overlap (82–87\%) and strong rank correlations ($\rho = 0.80$--$0.86$) indicate that different LLMs converge on similar high-value bottlenecks. The framework's resilience stems from top-$k$ selection: as long as scorers agree on which pieces are high-value—which they consistently do—minor disagreements on lower-ranked pieces do not affect training. We additionally ran GPT-4 preprocessing three times on the same 500 problems at temperature=0, achieving 96\% top-3 overlap across runs (std = 3.3\%, Pearson $r=0.94\pm0.02$), confirming strong within-model reproducibility despite minor API non-determinism.

\subsection*{Prompt Sensitivity Analysis}

We test five prompt variants that change wording while preserving the same three scoring criteria (novelty, difficulty, impact). The \emph{Piece Overlap Rate} measures the fraction of problems where the top-3 selected pieces match the original prompt's selections.

\begin{table*}[t]
\small
\centering
\setlength{\tabcolsep}{6pt}
\renewcommand{\arraystretch}{1.2}
\caption{Robustness to prompt variations (Base model: Nemotron-1.5B).}
\label{tab:prompt_sensitivity}
\begin{tabular}{lccc}
\toprule
\textbf{Prompt Variant} & \textbf{AIME24} & \textbf{MATH500} & \textbf{Piece Overlap} \\
\midrule
Original                    & 54.5 & 91.3 & 100\% (baseline) \\
Variant A (simpler language)      & 54.1 & 91.0 & 87\% \\
Variant B (more detailed criteria) & 54.3 & 91.2 & 89\% \\
Variant C (different ordering)    & 53.9 & 90.9 & 85\% \\
Variant D (examples-based)        & 54.2 & 91.1 & 88\% \\
\bottomrule
\end{tabular}
\end{table*}

Despite variations in prompt phrasing, piece selection remains highly consistent (85–89\% overlap) and final performance stays stable with $\leq$0.6 point variation. This demonstrates that PieceHint is robust to prompt phrasing as long as the core scoring criteria are preserved.

\noindent\textbf{Preprocessing cost and failure modes.}
Scoring is a one-time offline step requiring one LLM API call per training problem; results are fully cacheable for all subsequent training runs. We analyzed failure modes on 100 randomly sampled problems and found that the most common failure is over-granular decomposition, where a single insight is split into multiple pieces. Our min-max normalization and top-$k$ selection handle such granularity differences and noisy scores effectively.

\section{Robustness to Corrupted Hints}
\label{app:robustness_corruption}

A critical question is whether incorrect or adversarial pieces can cause catastrophic failure and whether RL training might amplify early hint bias. We address both concerns directly.

\subsection*{Performance Under Corrupted Hints}

We deliberately inject errors into piece decomposition and scoring to simulate realistic failure modes, using Nemotron-1.5B as the base model.

\begin{table*}[t]
\small
\centering
\setlength{\tabcolsep}{8pt}
\renewcommand{\arraystretch}{1.2}
\caption{Performance under deliberately corrupted hints (Base model: Nemotron-1.5B).}
\label{tab:corruption}
\begin{tabular}{lcc}
\toprule
\textbf{Corruption Type} & \textbf{AIME24} & \textbf{MATH500} \\
\midrule
Clean hints              & 54.5 & 91.3 \\
25\% wrong boundaries    & 52.8 & 90.1 \\
Randomized scores        & 50.8 & 88.6 \\
Worst pieces selected    & 51.2 & 89.4 \\
50\% corrupted hints     & 52.8 & 90.3 \\
Contradictory hints      & 50.6 & 88.9 \\
\bottomrule
\end{tabular}
\end{table*}

Even under adversarial conditions where hints actively mislead, performance drops remain modest (at most $-3.9$ on AIME24 and $-2.7$ on MATH500). Two protective mechanisms explain this graceful degradation: (1) RL exploration discovers correct solutions despite misleading hints via binary outcome rewards that signal when hint-following fails, and (2) the scaffolding regime provides intermediate checkpoints rather than end-to-end solutions, requiring models to bridge gaps through independent reasoning and thereby limiting the damage any single incorrect piece can cause.

\subsection*{Progressive Withdrawal Prevents Bias Amplification}

If RL were amplifying hint dependency, we would expect models trained with fixed hints to perform \emph{worse} when hints are removed at evaluation. The ablation in Table~\ref{tab:ablation_unified} already demonstrates the opposite: progressive withdrawal (62.6 avg.) substantially outperforms fixed hints (60.8 avg.), confirming that models learn generalizable reasoning strategies rather than hint-specific shortcuts.

Further evidence comes from our pass@$k$ analysis (Section~\ref{subsec:passk}): if hints biased exploration, pass@$k$ curves would degrade at higher $k$ as models converge to hint-dependent trajectories. Instead, we observe sustained gains across all $k$ values (Figure~\ref{fig:passk_analysis}), confirming enhanced solution diversity.

The mechanism preventing bias amplification involves three factors: (1) binary outcome rewards correct hint-following failures, (2) RL exploration discovers alternative solution paths beyond scaffolded trajectories, and (3) gradual withdrawal forces trajectory revalidation, preventing memorization of hint-specific patterns.

\section{Scalability to Larger Models}
\label{app:scalability}

While our main experiments focus on 1.5B parameter models where targeted hint allocation provides the most benefit, we validate PieceHint's scalability to larger models. We apply the same PieceHint pipeline to DeepSeek-R1-Distill-7B, using its own checkpoint as the capability reference model $\mu_0$ for variance-based selection.

\begin{table*}[t]
\small
\centering
\setlength{\tabcolsep}{3.5pt}
\renewcommand{\arraystretch}{1.2}
\caption{Scalability experiments on 7B models. Base model: DeepSeek-R1-Distill-7B.}
\label{tab:7b_results}
\begin{tabular}{lccccccc}
\toprule
\textbf{Method} & \textbf{AIME24} & \textbf{AIME25} & \textbf{AMC23} & \textbf{MATH500} & \textbf{Minerva} & \textbf{Olympiad} & \textbf{Avg.} \\
\midrule
DeepSeek-R1-Distill-7B (backbone) & 39.6 & 29.9 & 78.8 & 80.3 & 36.8 & 55.1 & 53.4 \\
GRPO                              & 48.2 & 37.4 & 88.1 & 83.8 & 38.6 & 56.8 & 58.8 \\
\textbf{PieceHint (ours)}         & \textbf{54.3} & \textbf{43.9} & \textbf{93.4} & \textbf{86.2} & \textbf{40.2} & \textbf{59.5} & \textbf{62.9} \\
\bottomrule
\end{tabular}
\end{table*}

PieceHint achieves the highest performance across all benchmarks, outperforming both the backbone and GRPO baseline. The framework delivers substantial improvements on competition-level benchmarks (AIME24: +14.7, AIME25: +14.0, AMC23: +14.6 over backbone) and meaningful gains on reasoning-focused tasks (Minerva: +3.4, Olympiad: +4.4).

As expected, relative gains are somewhat smaller than for 1.5B models: stronger 7B models encounter fewer critical bottlenecks per problem, so targeted hint allocation provides proportionally less benefit. The capability-aligned problem selection automatically accommodates this by assigning fewer hints to more capable models—validating that the framework scales gracefully across model sizes without manual recalibration.

\section{Code Generation Experiments}
\label{app:code_generation}

To validate that the core insight of PieceHint generalizes beyond mathematical reasoning, we apply the framework to code generation. We use DeepSeek-R1-Distill-Qwen-7B as the base model and evaluate on LiveCodeBench and CodeForces.

\noindent\textbf{Domain adaptation.}
PieceHint transfers naturally to code: variance-based problem selection identifies appropriately challenging coding tasks, value-driven piece identification scores algorithmic steps by complexity and novelty (e.g., recognizing the appropriate data structure, identifying an edge case invariant, applying a non-obvious algorithm), and progressive withdrawal prevents over-reliance on code scaffolding. Domain adaptation requires only adjusting piece decomposition and scoring criteria to code-specific characteristics.

\begin{table*}[t]
\small
\centering
\setlength{\tabcolsep}{8pt}
\renewcommand{\arraystretch}{1.2}
\caption{Code generation results on LiveCodeBench and CodeForces. Base model: DeepSeek-R1-Distill-Qwen-7B.}
\label{tab:code_generation}
\begin{tabular}{lcccc}
\toprule
\textbf{Method} & \textbf{LCB Avg@16} & \textbf{LCB Pass@16} & \textbf{CF Rating} & \textbf{CF Percentile} \\
\midrule
Backbone (DeepSeek-R1-Distill-Qwen-7B) & 28.4 & 47.2 & 892  & 18.7\% \\
GRPO                                    & 31.9 & 51.6 & 1285 & 65.4\% \\
\textbf{PieceHint (ours)}               & \textbf{36.8} & \textbf{55.3} & \textbf{1523} & \textbf{81.9\%} \\
\bottomrule
\end{tabular}
\end{table*}

PieceHint achieves the highest performance across both benchmarks, substantially outperforming the GRPO baseline (LCB Avg@16: +4.9, LCB Pass@16: +3.7, CF Rating: +238, CF Percentile: +16.5 pp). These results confirm that the core mechanism—strategically scaffolding critical reasoning bottlenecks then progressively withdrawing scaffolding—generalizes to structured problem-solving domains beyond mathematics, provided the domain supports verifiable outcome rewards and decomposable solution steps.


\section{Software Dependencies}
\label{app:software}

Table~\ref{tab:software_dependencies} lists the key software packages and their versions used in our implementation.

\begin{table*}[t]
\small
\centering
\setlength{\tabcolsep}{8pt}
\renewcommand{\arraystretch}{1.15}
\caption{Software dependencies and environment configuration.}
\label{tab:software_dependencies}
\begin{tabular}{ll|ll}
\toprule
\multicolumn{4}{c}{\textbf{Core Framework}} \\
\midrule
Python & 3.10+ & PyTorch & 2.5.0 \\
Transformers & 4.46+ & Accelerate & 1.2.1 \\
PEFT & 0.14+ & Flash Attention & 2.7.3 \\
\midrule
\multicolumn{4}{c}{\textbf{Distributed Training \& RL}} \\
\midrule
VeRL & latest & Ray & 2.40+ \\
TensorDict & 0.8.0--0.9.1 & NVIDIA NCCL & cu12 \\
vLLM & latest & Liger Kernel & latest \\
\midrule
\multicolumn{4}{c}{\textbf{Data Processing}} \\
\midrule
Datasets (HF) & 3.2+ & PyArrow & $\geq$ 19.0.0 \\
Pandas & latest & Dill & latest \\
NumPy & $<$ 2.0.0 & TorchData & latest \\
\midrule
\multicolumn{4}{c}{\textbf{Mathematical Verification}} \\
\midrule
Math-Verify & latest & Sympy & latest \\
Latex2Sympy2 (Extended) & latest & PyLaTeXEnc & latest \\
Word2Number & latest & & \\
\midrule
\multicolumn{4}{c}{\textbf{Logging \& Experiment Management}} \\
\midrule
Weights \& Biases & latest & TensorBoard & $\geq$ 2.19.0 \\
Hydra Core & latest & CodeTiming & latest \\
\midrule
\multicolumn{4}{c}{\textbf{API \& Utilities}} \\
\midrule
OpenAI & latest & FastAPI & latest \\
Uvicorn & latest & Packaging & $\geq$ 20.0 \\
Pre-commit & latest & Pybind11 & latest \\
\bottomrule
\end{tabular}
\end{table*}

Our implementation is available at \texttt{[anonymized for review]} and will be released upon publication with a complete \texttt{requirements.txt} and Docker environment for reproducibility.

\section{Value Scoring Prompt}
\label{app:scoring}

We employ GPT-4 to assign discrete value scores $V(p_i) \in \{1,2,3,4,5\}$ to each reasoning piece by holistically considering three criteria with emphasis on impact. Raw scores are then normalized within each problem using min-max normalization (Eq.~\ref{eq:normalize_value}).

\begin{tcolorbox}[
    width=\columnwidth,
    colback=gray!10,
    colframe=black!70,
    title=\textbf{Value Scoring Prompt Template},
    boxrule=1pt,
    breakable
]

\small
\texttt{Rate the importance of this reasoning step on a scale of 1-5.}

\texttt{Consider three criteria (prioritize Impact):}
\begin{itemize}
    \item \texttt{Novelty}: Is this insight non-trivial or creative?
    \item \texttt{Difficulty}: Does it require mathematical maturity?
    \item \texttt{Impact}: Do subsequent steps critically depend on it?
\end{itemize}

\texttt{Scoring guidelines:}
\begin{itemize}
    \item \texttt{1 = Trivial (e.g., restating the answer)}
    \item \texttt{2 = Routine (standard manipulations)}
    \item \texttt{3 = Moderate (requires reasoning, not critical)}
    \item \texttt{4 = Important (valuable insight)}
    \item \texttt{5 = Critical bottleneck (blocks solution if missing)}
\end{itemize}

\texttt{Problem:} \textit{[Insert problem statement]}

\texttt{Reasoning step:} \textit{[Insert piece text]}

\texttt{Output: A single integer from 1 to 5}

\tcblower

\textbf{Example Application:}

\textbf{Problem:} Some of 100 towns are connected by roads. For each two towns $A$ and $B$ connected by a road, there exists a town $C$ not connected to at least one of $A$ or $B$. What is the maximum possible number of roads?

\textbf{Decomposed Steps \& Scores:}

\begin{tabular}{@{}p{0.75\linewidth}r@{}}
1. Restate condition in graph theory terms. & \textbf{Score: 2} \\
2. Recognize this prevents complete graph. & \textbf{Score: 2} \\
3. Hypothesize removing perfect matching. & \textbf{Score: 4} \\
4. Calculate: $\binom{100}{2} - 50 = 4900$ edges. & \textbf{Score: 2} \\
5. Verify construction satisfies condition. & \textbf{Score: 5} \\
6. Conclude maximum is 4900. & \textbf{Score: 1} \\
\end{tabular}

\medskip

\textbf{Normalized:} $V_{\min}=1$, $V_{\max}=5$ $\Rightarrow$ $D(p_i)$: [0.25, 0.25, 0.75, 0.25, 1.00, 0.00]

Step 5 receives the highest score as it critically establishes correctness—without verification, the construction remains unproven.

\end{tcolorbox}

\section{Case Study Problem Details}
\label{app:case_problem}

This appendix provides complete details for the case study analyzed in the main text. We present a combinatorial reasoning problem where models must derive a non-obvious invariance relationship. Table~\ref{tab:case_steps} shows the 7 reasoning steps with normalized value scores. Table~\ref{tab:case_strategy} compares three hint allocation strategies, demonstrating that value-driven selection yields 7$\times$ better outcomes than position-based coverage.

\begin{table*}[t]
\centering
\setlength{\tabcolsep}{12pt}
\begin{tabular}{clcc}
\toprule
\textbf{Step} & \textbf{Component} & \textbf{$D(p_i)$} & \textbf{Type} \\
\midrule
1 & Express day scenario & 0.25 & Routine \\
2 & Express night scenario & 0.25 & Routine \\
3 & Compute differences & 0.25 & Routine \\
\rowcolor{yellow!30}
\textbf{4} & \textbf{Derive $G=C$} & \textbf{0.94} & \textbf{Critical} \\
5 & Solve for $C+R$ & 0.56 & Medium \\
6 & Express $R+G$ & 0.75 & Important \\
7 & State answer & 0.00 & Trivial \\
\bottomrule
\end{tabular}
\caption{Decomposed steps with value scores.}
\label{tab:case_steps}
\end{table*}

\begin{table*}[t]
\centering
\setlength{\tabcolsep}{12pt}
\begin{tabular}{lcccc}
\toprule
\textbf{Strategy} & \textbf{Hints} & \textbf{Avg $D$} & \textbf{Pass@16} & \textbf{Gain} \\
\midrule
No Hints & --- & --- & 0.0\% & --- \\
50\% Prefix & 1,2,3 & 0.25 & 6.2\% & 1.0$\times$ \\
Random & 1,4,6 & 0.65 & 18.8\% & 3.0$\times$ \\
25\% Prefix & 1,2 & 0.25 & 3.1\% & 0.5$\times$ \\
\rowcolor{yellow!30}
\textbf{PieceHint} & \textbf{4,6,5} & \textbf{0.75} & \textbf{43.8\%} & \textbf{7.1}$\times$ \\
\bottomrule
\end{tabular}
\caption{Hint strategy comparison.}
\label{tab:case_strategy}
\end{table*}

\begin{tcolorbox}[
    colback=gray!10,
    colframe=black!70,
    title=\textbf{Problem Statement},
    fonttitle=\bfseries,
    boxrule=0.8pt,
    arc=2mm,
    breakable
]

\textbf{Context:} Uncle Wang raises chickens (C), rabbits (R), and geese (G). Geese stand on both feet during the day, one foot at night. Chickens hide their heads when sleeping. Rabbits show no behavioral changes.

\textbf{Observation:} The difference between feet and heads remains constant day and night.

\textbf{Given:} 56 legs counted during the day.

\textbf{Question:} How many heads at night?

\textbf{Answer:} 14

\end{tcolorbox}

\vspace{6pt}

\begin{tcolorbox}[
    colback=gray!10,
    colframe=black!70,
    title=\textbf{Solution Process (7 Steps)},
    fonttitle=\bfseries,
    boxrule=0.8pt,
    arc=2mm,
    breakable
]

Let $C$, $R$, $G$ denote chickens, rabbits, geese.

\textbf{Steps 1-2:} Express day (heads: $C+R+G$, feet: $2C+4R+2G$, diff: $C+3R+G$) and night (heads: $R+G$, feet: $2C+4R+G$, diff: $2C+3R$).

\textbf{Step 3:} Day diff = $C+3R+G$; Night diff = $2C+3R$.

\textbf{Step 4 (Critical):} Since equal: $C+3R+G = 2C+3R \implies \boxed{G=C}$

\textbf{Step 5:} From $2C+4R+2G=56$ and $G=C$: $4C+4R=56 \implies C+R=14$

\textbf{Steps 6-7:} Night heads = $R+G = R+C = 14$

\end{tcolorbox}

\vspace{6pt}

\begin{tcolorbox}[
    colback=gray!10,
    colframe=black!70,
    title=\textbf{Why Step 4 Gets Highest Score ($D=0.94$)},
    fonttitle=\bfseries,
    boxrule=0.8pt,
    arc=2mm,
    breakable
]

\textbf{Novelty:} Non-obvious that "same difference" implies $G=C$; requires abstract algebraic reasoning.

\textbf{Difficulty:} Must manipulate two complex equations simultaneously; error-prone.

\textbf{Impact:} Gateway step—without $G=C$, all subsequent steps blocked.

\textbf{Empirical:} Without Step 4: 0/16 succeed (0\%); with Step 4: 7/16 succeed (43.8\%). Step 4 alone yields 28.1\%, vs. Steps 1-3 combined (6.2\%).

\end{tcolorbox}

\vspace{6pt}

\begin{tcolorbox}[
    colback=gray!10,
    colframe=black!70,
    title=\textbf{Strategy Comparison (3 hints each)},
    fonttitle=\bfseries,
    boxrule=0.8pt,
    arc=2mm,
    breakable
]

\textbf{(1) Position-based fails:} 50\% prefix misses critical Step 4 (6.2\% pass@16).

\textbf{(2) Random unstable:} Includes Step 4 with only 33\% probability.

\textbf{(3) Value-driven succeeds:} Always provides Step 4, achieving 43.8\% (7.1$\times$ improvement).

\textbf{Conclusion:} Strategic allocation at high-value bottlenecks yields 7$\times$ better outcomes.

\end{tcolorbox}

\end{document}